\documentclass[manuscript,screen,review, acmsmall]{acmart}

\AtBeginDocument{%
  \providecommand\BibTeX{{%
    \normalfont B\kern-0.5em{\scshape i\kern-0.25em b}\kern-0.8em\TeX}}}

\usepackage[fleqn]{mathtools} 
\usepackage{amsfonts}
\usepackage{amsthm}
\usepackage[acronym]{glossaries}
\usepackage[useregional]{datetime2}
\usepackage{hyperref}
\usepackage{xspace}
\usepackage{enumitem}
\usepackage{tikz, subcaption}
\usetikzlibrary{arrows.meta}
\usepackage{amsmath}
\usepackage{geometry, pdflscape, ltablex, ragged2e, multirow,booktabs}
\usepackage{tikz}
\usepackage{booktabs}
\usepackage{multirow}
\usetikzlibrary{shapes.geometric, arrows.meta, positioning, calc}

\usepackage{pgfplots} 
\usepackage{pgfplotstable}
\pgfplotsset{width=15.5cm, height=6.5cm, compat=1.18}  

\usepackage{tabularx}
\usepackage{float}
\usepackage{parskip}
\graphicspath{{figures}}

\definecolor{paleyellow}{HTML}{FFE3C0}
\long\def\comment[#1]#2{\par\colorbox{paleyellow}{\llap{#1:\quad}%
    \parbox[t]{\textwidth}{\setlength{\parskip}{1ex plus 0.2ex minus 0.2ex}#2}}}

\pgfplotstableread{
Y	PosErr	NegErr
75.0	2.0	3.0
2.7	1.4	1.7
2.4	1.7	1.4
}\datatableisreasoningZero

\pgfplotstableread{
Y	PosErr	NegErr
100.0	0.0	0.0
100.0	0.0	0.0
100.0	0.0	0.0
}\datatableisreasoningOne

\pgfplotstableread{
Y	PosErr	NegErr
100.0	0.0	0.0
100.0	0.0	0.0
100.0	0.0	0.0
}\datatableisreasoningTwo

\pgfplotstableread{
Y	PosErr	NegErr
100.0	0.0	0.0
97.0	1.0	1.0
100.0	0.0	0.0
}\datatableisreasoningThree

\pgfplotstableread{
Y	PosErr	NegErr
100.0	0.0	0.0
100.0	0.0	0.0
100.0	0.0	0.0
}\datatableisreasoningFour

\pgfplotstableread{
Y	PosErr	NegErr
100.0	0.0	0.0
100.0	0.0	0.0
100.0	0.0	0.0
}\datatableisreasoningFive

\pgfplotstableread{
Y	PosErr	NegErr
1.3	2.7	1.3
0.3	0.7	0.3
0.0	0.0	0.0
}\datatablehasstepsZero

\pgfplotstableread{
Y	PosErr	NegErr
1.0	1.9	1.0
46.3	35.7	32.3
23.1	41.1	21.0
}\datatablehasstepsOne

\pgfplotstableread{
Y	PosErr	NegErr
42.0	44.0	25.0
58.7	29.3	28.7
36.0	21.0	24.0
}\datatablehasstepsTwo

\pgfplotstableread{
Y	PosErr	NegErr
48.7	26.3	19.7
96.0	1.0	2.0
49.6	8.4	13.9
}\datatablehasstepsThree

\pgfplotstableread{
Y	PosErr	NegErr
25.3	38.7	20.3
86.7	7.3	14.7
32.3	30.7	22.3
}\datatablehasstepsFour

\pgfplotstableread{
Y	PosErr	NegErr
58.0	27.0	31.0
86.0	14.0	11.0
43.0	37.0	34.0
}\datatablehasstepsFive

\pgfplotstableread{
Y	PosErr	NegErr
0.3	0.7	0.3
0.3	0.7	0.3
0.0	0.0	0.0
}\datatableratiorightmethodandhasstepsZero

\pgfplotstableread{
Y	PosErr	NegErr
0.3	0.7	0.3
1.0	0.0	0.0
1.0	0.0	0.0
}\datatableratiorightmethodandhasstepsOne

\pgfplotstableread{
Y	PosErr	NegErr
1.0	0.0	0.0
1.0	0.0	0.0
1.0	0.0	0.0
}\datatableratiorightmethodandhasstepsTwo

\pgfplotstableread{
Y	PosErr	NegErr
0.9	0.0	0.0
0.7	0.1	0.1
0.5	0.1	0.2
}\datatableratiorightmethodandhasstepsThree

\pgfplotstableread{
Y	PosErr	NegErr
0.4	0.2	0.3
0.2	0.3	0.2
0.2	0.3	0.2
}\datatableratiorightmethodandhasstepsFour

\pgfplotstableread{
Y	PosErr	NegErr
0.1	0.2	0.1
0.2	0.3	0.2
-0.0	0.0	0.0
}\datatableratiorightmethodandhasstepsFive

\newcolumntype{L}{>{$}l<{$}}%
\newcolumntype{C}{>{$}c<{$}}%
\newcolumntype{R}{>{$}r<{$}}%
\newcolumntype{M}{@{}p{\mathindent}@{}}%
\newcolumntype{t}{>{\mbox\bgroup}l<{\egroup}}%
\newcolumntype{e}{r@{\;}l}%
\newcolumntype{E}{>{$}r<{$}@{$\;$}>{$}l<{$}}%

\makeatletter
\def\@ACM@badge@width{0pt}
\def\@ACM@badge@skip{0pt}
\def\@mkteasers{}
\makeatother

\begin{document}

\title{Steamroller Problems: An Evaluation of LLM Reasoning Capability with Automated Theorem Prover Strategies}

\author{Lachlan McGinness}
\email{lachlan.mcginness@anu.edu.au}
\affiliation{%
  \institution{School of Computer Science, Australian National University and CSIRO}
}
\author{Peter Baumgartner}
\email{peter.baumgartner@data61.csiro.au}
\affiliation{%
  \institution{Data61, CSIRO}
}

\renewcommand{\shortauthors}{McGinness, L. and Baumgartner, P.}

\begin{abstract}

This study presents the first examination of the ability of Large Language Models (LLMs) to follow reasoning strategies that are used to guide Automated Theorem Provers (ATPs). We evaluate the performance of GPT4, GPT3.5 Turbo and Google's recent Gemini model on problems from a steamroller domain. In addition to determining accuracy we make use of the Natural Language Processing library spaCy to explore new methods of investigating LLM's reasoning capabilities. This led to one alarming result, the low correlation between correct reasoning and correct answers for any of the tested models.  We found that the models' performance when using the ATP reasoning strategies was comparable to one-shot chain of thought and observe that attention to uncertainty in the accuracy results is critical when drawing conclusions about model performance. Consistent with previous speculation we confirm that LLMs have a preference for, and are best able to follow, bottom up reasoning processes. However, the reasoning strategies can still be beneficial for deriving small and relevant sets of formulas for external processing by a trusted inference engine.
\end{abstract}




\maketitle

\section{Introduction}
\label{sec:Introduction}

Despite incrementally improving over the last several decades \cite{Khurana2023Natural}, Natural Language Processing (NLP) took the world by a storm in early 2023 with the release of GPT3 through Chat-GPT \cite{Brown2020Language}. Since then there has been significant hype around Large Language Models (LLMs) and many companies embedding them into their applications \cite{Zhao2023Survey} and many people are speculating about the use of LLMs in fields such as law, healthcare and education \cite{Bommasani2022Opportunities}. One of the main causes of excitement are the emergent properties of LLMs, where they seem to be able to store and recall knowledge and even logically reason. 

There are a number of benchmarks which have been designed to evaluate Large Language Models \cite{Zhao2023Survey}. A number of these tasks are strictly in the NPL domain \cite{Bajor2014Findings, Denis2016Lambada, Minaee2021Deep, Socher2013Recursive, Yang2018HotptQA}. However a number of these tasks have also been designed to evaluate emergent properties such as mathematical reasoning, coding or commonsense reasoning \cite{Austin2021Program, Chen2021Evaluating, Cobbe2021Training, Hendrycks2021Measuring, Zellers2019HellaSwag, McGinness2024Automated}. 

There exist many LLM leaderborads and there has been a strong focus on beating current state of the art performance on these benchmarks, even by very narrow margins. In these scenarios there is often no consideration of uncertainty in the presented scores. Also in logical reasoning benchmarks there is often a focus on the correct answers rather than the correct steps in reasoning. Often the computational expense is not explicitly mentioned for LLM performance on logical reasoning benchmarks. 

In this paper we evaluate three of the highest performing Large Language Models of December 2023 on PRONTOQA, a logical reasoning benchmark \cite{Saparov2023Language}. The models are GPT3.5 Turbo (which we will refer to as GPT3 from here on), GPT-4 and Google's Gemini-Pro. GPT3 and GPT4 were accessed through Microsoft Azure endpoints. Unlike previous studies, we use prompts to `teach' LLMs to use reasoning strategies used to control Automated Theorem Provers (ATPs) and related systems. Such strategies have been devised with reasoning correctness and search space pruning in mind. The underlying research question is whether LLM reasoning can benefit from these strategies. We not only report on the accuracy of the models in such settings, we provide uncertainty values reflecting the range of model performance in our experiments. Additionally we evaluate the reasoning of the models with uncertainty,  verify that their reasoning process contains all required steps and evaluate how faithfully the models follow the ATP reasoning algorithms. 

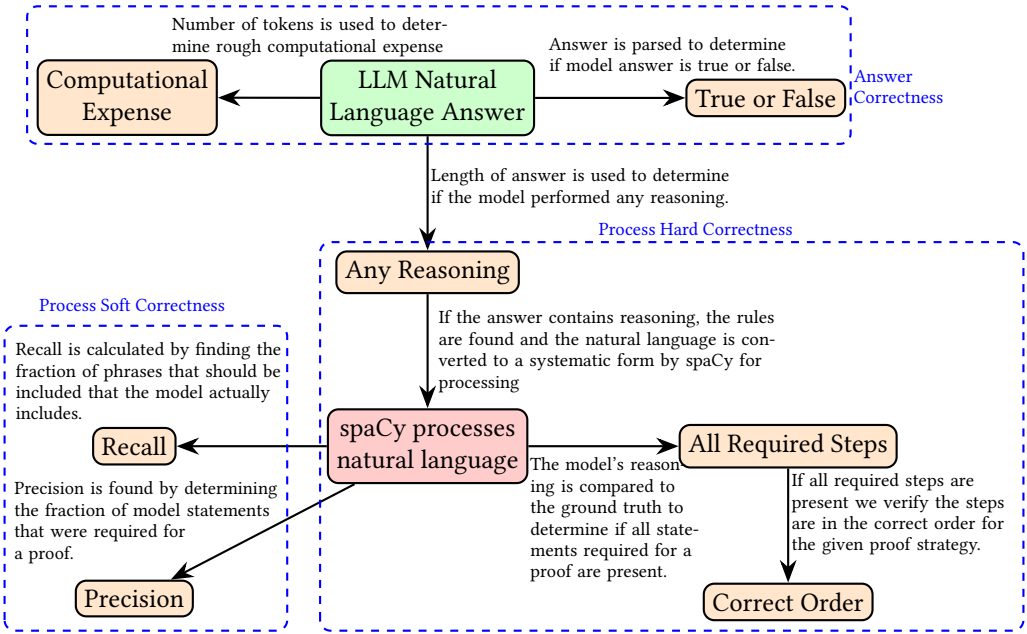
\begin{figure}[h!]
    \centering
    \begin{tikzpicture}[node distance=1.5cm and 2.0cm,
    box/.style={rounded corners, draw, align=center, thick, fill=#1!20},
    arrow/.style={-{Stealth[length=3mm]}, thick},
    arrowlabel/.style={font=\scriptsize, fill=none, inner sep=1pt}]

    \node[box=green] (answer) {LLM Natural\\ Language Answer};
    \node[box=orange, right=of answer] (accuracy) {True or False};
    \node[box=orange, below=of answer] (contains) {Any Reasoning};
    \node[box=orange] (tokens) at ($(accuracy.north) + (-8.5,-0.25)$) {Computational \\ Expense};
    \node[box=red, below=of contains] (spacy) {spaCy processes \\ natural language};
    \node[box=orange, right=of spacy] (required) {All Required Steps};
    \node[box=orange, below=of required] (order) {Correct Order};
    \node[box=orange, left=of spacy] (recall) {Recall};
    \node[box=orange, below=of recall] (precision) {Precision};

    \draw[arrow] (answer) -- (accuracy) node[arrowlabel, pos=1.02, above=0.3, align=left, text width=3.7cm] {Answer is parsed to determine \\if model answer is true or false.};
    \draw[arrow] (answer) -- (contains) node[arrowlabel, pos=0.45, right=0.0, text width=4.2cm] {Length of answer is used to determine \\if the model performed any reasoning.};
    \draw[arrow] (answer) -- (tokens) node[arrowlabel, pos=0.05, above=0.5, text width = 3.8cm] {Number of tokens is used to determine rough computational expense};
    \draw[arrow] (contains) -- (spacy) node[arrowlabel, pos=0.5, right=0.1, text width=4.5cm] {If the answer contains reasoning, the rules are found and the natural language is converted to a systematic form by spaCy for processing};
    \draw[arrow] (spacy) -- (required) node[arrowlabel, pos=0.64, below=0.1, text width = 2.5cm] {The model's reason-\\ing is compared to \\the ground truth to determine if all statements required for a proof are present.};
    \draw[arrow] (required) -- (order) node[arrowlabel, pos=0.4, right, text width=3.0cm] {If all required steps are present we verify the steps are in the correct order for the given proof strategy.};
    \draw[arrow] (spacy) -- (recall) node[arrowlabel, pos=1.12, above=0.3, text width=3.8cm] {Recall is calculated by finding the fraction of phrases that should be included that the model actually includes.};
    \draw[arrow] (spacy) -- (precision) node[arrowlabel, pos=0.4, left, text width=3.5cm] {Precision is found by determining the fraction of model statements that were required for \\a proof.};

    \draw[blue, dashed, thick, rounded corners] ($(answer.south west)+(-3.9,-0.1)$) rectangle ($(answer.north east)+(4.2,0.7)$);
    \node[align=left, font=\scriptsize, blue, anchor=north] at ($(answer.south west)+(7.7,1)$) (caption) {Answer\\ Correctness};
    \draw[blue, dashed, thick, rounded corners] ($(answer.south west)+(-4.2,-6.5)$) rectangle ($(answer.north east)+(-3.3,-3.5)$);
    \node[align=left, font=\scriptsize, blue, anchor=north] at ($(answer.south west)+(-2.5,-2)$) (caption) {Process Soft Correctness};
    \draw[blue, dashed, thick, rounded corners] ($(answer.south west)+(0,-6.5)$) rectangle ($(answer.north east)+(6.5,-2.4)$);
    \node[align=left, font=\scriptsize, blue, anchor=north] at ($(answer.south west)+(5.0,-1)$) (caption) {Process Hard Correctness};
    
\end{tikzpicture}
    \caption{Diagram of analysis pipeline. This shows all of the quantities that were investigated in this study. The Answer Correctness blue box corresponds to the traditional analysis seen in most studies using benchmarks. Process Hard Correctness evaluates qualities of the model's reasoning process using a Boolean framework. Process Soft Correctness evaluates the model's reasoning by considering the extent of completeness and conciseness of responses.}
    \label{fig:analysisPipeline}
\end{figure}

\section{Background}
\label{sec:background}

\subsection{LLM reasoning}
\label{sec:LLMreasoning}

Although Large Language Models were originally developed for next token prediction, they have been shown to possess a number of emergent properties including arithmetic, question answering and forms of complex reasoning \cite{Wei2022Emergent}. This is a fast moving field with continual release of more powerful models and more difficult benchmarks. The field is moving so quickly that on average less than two years pass from the time a benchmark is released until it is redundant because models have reached super-human or near perfect performance \cite{Srivastava2023Beyond}. 

In a summary paper Chang et al. suggest four general criteria for LLM performance that should apply across domains; accuracy, calibration, fairness and robustness \cite{Chang2024}. When evaluating LLM reasoning, most authors focus on the first of these criteria, accuracy \cite{Poesia2023Certified, binz_using_2023}. However obtaining the correct answer does not necessarily imply that models have used sound reasoning. Models may have seen the exact question and answer within their training data (contamination), have sufficient stored knowledge to answer the question without reasoning or simply guess the correct answer. 

In 2023, Saparov and He published a new dataset, PRONTOQA, specifically designed to allow for assessment of LLM's reasoning process rather than just their accuracy in completing the task \cite{Saparov2023Language}. This dataset requires models to use deductive reasoning (modus ponens) to answer a true or false query. The dataset is flexible with a number of parameters that can be changed including the number of steps of reasoning (hops), the ontology (whether the provided statements are true or false in the real world) and distractors (extra statements that are not required to answer the question). One added bonus of PRONTOQA is that the code which generates the questions is published, but the questions themselves are not. This decreases the chance of contamination and means that arbitrarily many examples can be generated. 

In addition to releasing the dataset Saparov and He also evaluate some of the easier problems in the data using the GPT3 model \cite{Saparov2023Language}. Specifically they investigate whether the accuracy of models correlates with their ability to generate a strict proof. They discovered that there is a weak correlation between accuracy and perfect (`golden' chain of thought) reasoning. However there is a much stronger correlation between accuracy and two weaker forms of proofs; proofs which are allowed to skip steps and proofs which only include valid steps \cite{Saparov2023Language}. They conclude that the longer proofs in PRONTOQA are still challenging for Large Language models and notice that the information in which the facts and rules are displayed has an impact on Model reasoning \cite{Saparov2023Language}.

In 2023, Xu et al. introduced a system for classifying the types of errors that LLMs make in their reasoning process \cite{Xu2023Large}. They break the errors into two categories, which contain error sub-types. The first category is an evidence selection error, where a model does not choose the correct initial statements. This is a common problem in Retrieval Augmented Generation (RAG) especially when the desired information is contained in the middle section of a long paragraph of text \cite{Liu2024Lost}. There are two sub-types of evidence selection errors; hallucination and wrong selection. In our work the model is only provided with a short context window and instead the second error category, reasoning process, is more relevant. Xu et al. divide this into three error types: (1) No reasoning which occur in $19.33\%$ of cases, (2) perspective mistake in $44.47\%$ of cases and (3) process mistake in $36.20\%$, but they do not explicitly define the difference between a perspective and process mistake \cite{Xu2023Large}. 

To improve performance of a pre-trained model on a reasoning task there are a number of methods that can be employed. Fine tuning is a technique which can increase a model's performance on a very specific task if there are a significant number of training examples to learn from \cite{Radford2019Language}. Fine tuning updates the weights of the model and so requires a significant amount of computational power, however it is commonly used to improve performance on benchmarks \cite{Ahn2024Large, Bisk2020PIQA, Touvron2023Llama2, Yue2023DISCLAWLLM}. Another method that can be used is prompt engineering where a human (or computer \cite{Shin2020Autoprompt, Shum2023Automatic}) phrases the task in such a way that it will increase the model's performance \cite{Brown2020Language, Kojima2022Large, Nye2022Show}. Most prompt engineering methods require very few or no examples and no extra computational power for additional training. Chen et al. and Liu et al. provide an overview of the existing prompt engineering techniques \cite{Chen2023Unleashing, Liu2023Pre-train}. 

The most widely used prompt engineering technique is Chain of Thought \cite{Chen2023Unleashing}, where models tend to use bottom up (or forward chain) reasoning \cite{Kazemi2023LAMBADA}. Kaxemi et al. from the Google Research team suggest a different approach where they implement top down (also known as goal oriented or backward chain) reasoning \cite{Kazemi2023LAMBADA}. Inspired by the success of backward chaining in the area of Automated Theorem Proving (ATP), they designed an algorithm called LAMBADA which makes multiple calls to an LLM to complete four different tasks 1.) Fact Checking, 2.) Rule Selection, 3.) Goal Decomposition and 4.) Sign Agreement. They demonstrate a significant performance improvement over Chain of Thought and Selection Inference techniques \cite{Kazemi2023LAMBADA}. We extend this work, rather than using an external algorithm which calls a Large Language model for different sub-tasks, we explicitly `teach' an LLM to use reasoning techniques for bottom up, top down and for bottom up guided in the spirit of the magic set transformation \cite{Bancilhon1985Magic}.

\subsection{LLM Computational Cost}
Besides the performance of Large Language models, their computational cost, power consumption and environmental cost has also been discussed \cite{Bender2021StochasticParrots, Faiz2024LLMCarbon}. Although there are methods to measure the computational expense of running an LLM task, it is rare for LLM users who are conducting experiments on their models to report the computational cost at inference time \cite{Chang2024, Naveed2024Comprehensive}. This stands in contrast to many other areas in computer science where understanding (and accurately reporting) the computational cost of running an algorithm is critical for determining their scalability, practicality and for further development and improvement. 

A number of theoretical and empirical ways of measuring the cost of algorithms are well established including CPU time, clock cycles, time complexity analysis and space complexity analysis. For Large Language Models there are well established metrics for training including Floating Point Operations (FLOPs), parameter counts, training time and energy consumption. For open source models these metrics are usually published, however not all metrics are made public for state of the art closed models \cite{Google2023Gemini, Openai2023GPT4}. During inference, the metrics most commonly discussed are those relating to performance such as latency and throughput. In this paper, we suggest and implement a simple measure for the rough computational expense of calling an LLM; the number of completion tokens. This allows for a direct comparison of computational resources used when comparing different techniques and tasks for a given model. This however does not allow for a fair comparison between different models as models can have different numbers of parameters. 

\subsection{ATP Reasoning Strategies}

\begin{figure}[h!]
    \centering
    \begin{minipage}[b]{0.45\linewidth}
    \begin{tikzpicture}[node distance=1cm and 1.5cm,
    box/.style={rounded corners, draw, align=center, thick, fill=#1!20},
    arrow/.style={-{Stealth[length=3mm]}, thick},
    arrowlabel/.style={font=\scriptsize, fill=none, inner sep=1pt}]

    \node[box=green] (query) {True or false: Whiskers is an animal.};
    \node[box=orange, below=of query] (rule1) {Every mammal is an animal.};
    \node[box=orange, below=of rule1] (rule2) {Each cat is a mammal.};
    \node[box=red, below=of rule2] (fact) {Whiskers is a cat.};
    \node (reasoningmethod) at ($(query.north) + (0,0.5cm)$) {Bottom Up Reasoning};
    
    \draw[arrow] (rule1) -- (query) node[arrowlabel, pos=0.35, right, align=left, text width=2.2cm] {Therefore Whiskers is an animal};
    \draw[arrow] (rule2) -- (rule1) node[arrowlabel, pos=0.35, right, align=left, text width=2.2cm] {Therefore Whiskers is a mammal};
    \draw[arrow] (fact) -- (rule2) node[arrowlabel, pos=0.5, above, align=center, text width=2.2cm] {};
    
\end{tikzpicture}
\end{minipage}
\begin{minipage}[b]{0.45\linewidth}
    \begin{tikzpicture}[node distance=1cm and 1.5cm,
    box/.style={rounded corners, draw, align=center, thick, fill=#1!20},
    arrow/.style={-{Stealth[length=3mm]}, thick},
    arrowlabel/.style={font=\scriptsize, fill=none, inner sep=1pt}]

    \node[box=green] (query) {True or false: Whiskers is an animal.};
    \node[box=orange, below=of query] (rule1) {Every mammal is an animal.};
    \node[box=orange, below=of rule1] (rule2) {Each cat is a mammal.};
    \node[box=red, below=of rule2] (fact) {Whiskers is a cat.};
    \node (reasoningmethod) at ($(query.north) + (0,0.5cm)$) {Top Down Reasoning};
        
    \draw[arrow] (query) -- (rule1) node[arrowlabel, pos=0.5, above, align=left, text width=2.2cm] {};
    \draw[arrow] (rule1) -- (rule2) node[arrowlabel, pos=0.35, right, align=left, text width=2.5cm] {True or False: Whiskers is a mammal};
    \draw[arrow] (rule2) -- (fact) node[arrowlabel, pos=0.35, right, align=left, text width=2.5cm] {True or False: Whiskers is a cat};
    
\end{tikzpicture}
\end{minipage}
    \\
\hspace{0.5cm}

    \begin{minipage}[b]{0.9\linewidth}
    \begin{tikzpicture}[node distance=1cm and 2.0cm,
    box/.style={rounded corners, draw, align=center, thick, fill=#1!20},
    arrow/.style={-{Stealth[length=3mm]}, thick},
    arrowlabel/.style={font=\scriptsize, fill=none, inner sep=1pt}]

    \node[box=green] (query) {True or false:\\ Whiskers is an animal.};
    \node[box=orange, below=of query] (rule1) {Every mammal is an animal.};
    \node[box=orange, below=of rule1] (rule2) {Each cat is a mammal.};
    \node[box=red, below=of rule2] (fact) {Whiskers is a cat.};
    \node (reasoningmethod) at ($(query.north) + (0,0.5cm)$) {Magic Set Reasoning};
    \node[box=red, left=of fact] (factir1) {Whiskers is furry.};
    \node[box=red, right=of fact] (factir2) {Rex is a dog.};
    \node[box=orange, above=of factir2] (rule3) {Dogs are mammals.};

    \draw[arrow] (rule1) -- (query) node[arrowlabel, pos=0.35, right, align=left, text width=2.2cm] {Therefore Whiskers\\ is an animal};
    \draw[arrow] (rule2) -- (rule1) node[arrowlabel, pos=0.35, right, align=left, text width=2.2cm] {Therefore Whiskers\\ is a mammal};
    \draw[arrow] (fact) -- (rule2) node[arrowlabel, pos=0.5, above, align=center, text width=2.2cm] {};
    
    \draw[blue, dashed, thick, rounded corners] ($(fact.south west)+(-1,5.7)$) rectangle ($(fact.north east)+(1,-0.7)$);
    \node[align=left, font=\scriptsize, blue, anchor=north] at ($(fact.south west)+(5.90,5)$) (caption) {First the model uses approximative\\ top down reasoning to identify\\ relevant facts and rules. The remaining \\ facts and rules can then be ignored. };
    \node[align=left, font=\scriptsize, black, anchor=north] at ($(fact.south west)+(5.6,3.8)$) (caption) {Then the model performs bottom \\up reasoning on the reduced set \\of axioms.};
    
    \draw[-{Stealth[length=3mm]}, blue, thick] (-2.8,0) -- (-2.8,-5);

    \coordinate (cross1center) at (-4.8,-4.8); 
    \coordinate (cross2center) at (4.5,-4.8); 
    \coordinate (cross3center) at (4.5,-3.2);

    \draw[blue, thick] ($(cross1center)+(-1,-0.5)$) -- ($(cross1center)+(1,0.5)$); 
    \draw[blue, thick] ($(cross1center)+(-1,0.5)$) -- ($(cross1center)+(1,-0.5)$);
    
    \draw[blue, thick] ($(cross2center)+(-1,-0.5)$) -- ($(cross2center)+(1,0.5)$); 
    \draw[blue, thick] ($(cross2center)+(-1,0.5)$) -- ($(cross2center)+(1,-0.5)$);
    
    \draw[blue, thick] ($(cross3center)+(-1,-0.5)$) -- ($(cross3center)+(1,0.5)$); 
    \draw[blue, thick] ($(cross3center)+(-1,0.5)$) -- ($(cross3center)+(1,-0.5)$);

\end{tikzpicture}
\end{minipage}
    \caption{Illustration of Reasoning Strategies. Facts are shown in red, rules in yellow and queries in green. For simplicity only very simple chains of reasoning with few or no distractors are shown. Magic Set Transformation has been used in Automated Theorem Proving and related disciplines for search space pruning. }
    \label{fig:reasoning}
\end{figure}
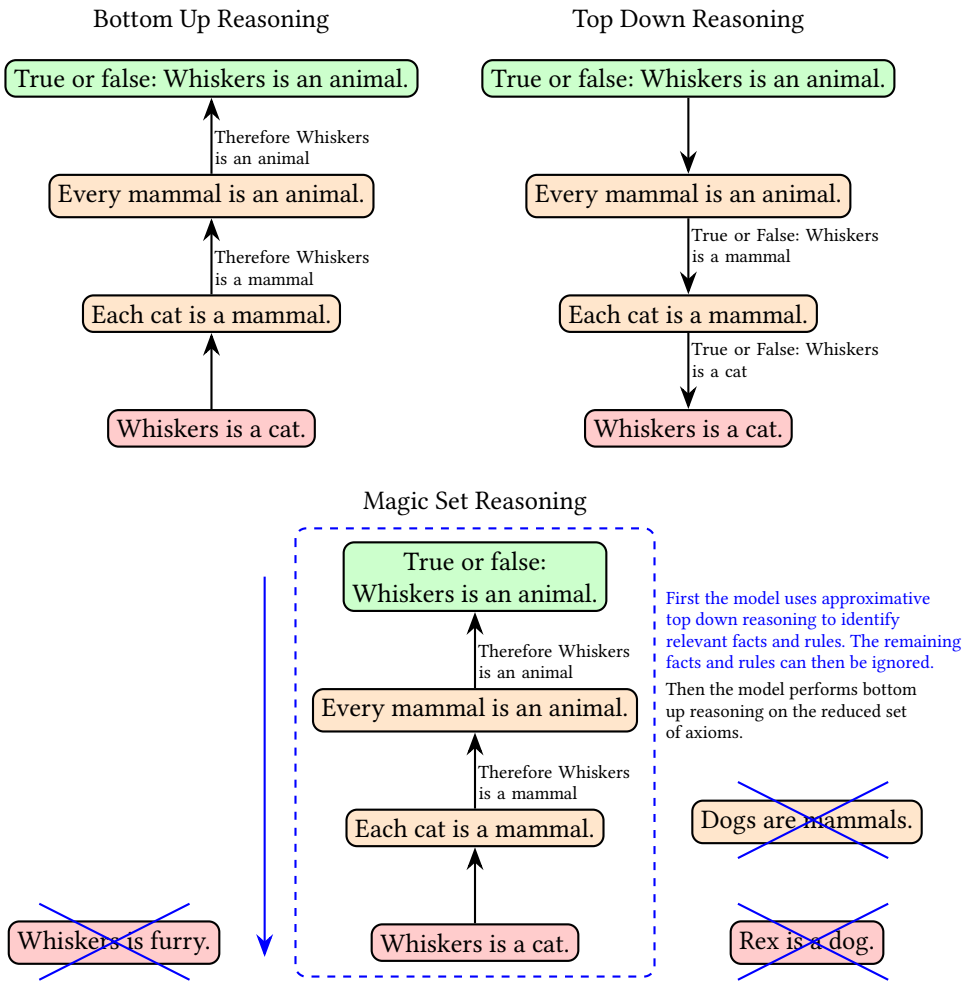

Two concepts that are routinely used in First Order Logic (FOL) Automated Theorem Prover (ATP) systems are bottom up and top down reasoning, also referred to as forward chaining and backward chaining respectively \cite{Harrison2009Handbook, Kazemi2023LAMBADA}. Bottom up begins a logical deduction process from basic facts and rules and iteratively derives conclusions until it has arrived at the answer to a query. By contrast top down reasoning systems start with a query and use rules to recursively derive sub-goals until the sub-goals and query can be proved or disproved by the provided facts. Top down reasoning has been used for guiding the search for a proof starting from the query. Bottom up is the primary form of reasoning used by LLMs when performing Chain of Thought (CoT) reasoning \cite{Kazemi2023LAMBADA}.

As mentioned in Section \ref{sec:LLMreasoning}, in 2023, Kazemi et al. from the GOOGLE research team were the first to attempt to use a top down approach when reasoning with LLMs \cite{Kazemi2023LAMBADA}. To do this they created an algorithm, LAMBADA which calls an LLM at various stages. As part of the study they perform a short preliminary experiment to determine if an LLM can perform CoT reasoning on statements where the proofs are written backwards but stop short of explicitly teaching an LLM to use top down reasoning. 

In addition to top down and bottom up reasoning we also explore the use of a magic set transformation approach with LLMs \cite{Bancilhon1985Magic}. Magic set transformation approaches first use a top down exploration to determine the set of rules and facts which are relevant to a query. This allows the subsequent bottom up reasoning to explore a smaller search space.

\section{Methods}
\label{sec:methods}

\subsection{Experiment Design}
For our experiments we choose to use problems from the popular steamroller domain. Steamroller problems are toy examples used by philosophers to test their mental capacity and logical reasoning abilities.  We choose to build on the work done by Saparov and He using the PRONTOQA dataset \cite{Saparov2023Language}. We extend their work by specifically evaluating the most current state of the art models' reasoning capabilities \cite{Brown2020Language, Google2023Gemini, Openai2023GPT4}. We accessed the GPT models using a Microsoft Azure API and accessed Gemini using a public API.

We focus not just on accuracy but also the correctness of their reasoning processes. In order to challenge the models we explore the False Ontology with distractors, the hardest conditions for reasoning, which were not tested in their original paper \cite{Dasgupta2023Language, Saparov2023Language}. 

For our baseline we take the score that would be expected what would we get from random guessing. This can be determined by sampling the binomial distribution $B(100, 0.5)$. To find the uncertainty we first sample the distribution three times and take the half range of these values. Then we repeat this process 10,000 times and take an average. The large Language Models were given six different queries corresponding to different techniques, in order to determine their reasoning ability:
\begin{itemize}
    \item Normal - The LLM is given just the question but no further instructions.
    \item Zero-shot CoT - The LLM is instructed to break the problem down into smaller steps and explain its reasoning. 
    \item One-shot CoT - The model is instructed to provide its reasoning and given an example of how to reason through the problem (using bottom up technique).
    \item Bottom Up - The model is given explicit instructions for how to perform bottom up reasoning and provided with an example. 
    \item Top Down - The model is given explicit instructions for how to perform top down reasoning and provided with an example. 
    \item Magic Set Transformation - The model is given explicit instructions for how to perform magic set transformation reasoning and provided with an example. 
\end{itemize}
We tested between two and ten prompts for each of these conditions. The prompt which yielded the highest accuracy was kept. Appendix \ref{sec:prompts} contains the exact prompts used for each of these conditions.

\paragraph{Answer correctness.}
For each experimental condition, we first determine whether the model correctly answered the query. In measuring accuracy we assume default negation; if the model refused to give an answer the question, or stated that the query could not be answered then its answer is taken as false. The number of correct answers that the model gave were used to calculate an accuracy score. 

\paragraph{Process hard correctness.}
In addition to determining the accuracy of the model for each technique, we also investigate whether the model used correct reasoning to answer the questions. Even when prompted to use reasoning, models occasionally simply answer True or False, which means that no reasoning has been used. Furthermore we investigate whether the model showed all of the required steps that would be needed by a human or ATP to prove the answer to the query. In addition to these we also specifically look for whether the models were able to follow the prescribed process, as outlined in Section \ref{sec:MethodsErrorTypes}.

\paragraph{Process soft correctness.}
Furthermore, we investigated a softer version of process correctness. Process soft correctness also considers the facts and rules in the model
response and compares them to ground truth. However,
the comparison is set-based, i.e., it ignores the order in which the statements appear.
From this we determine the degrees of completeness and conciseness in the model response compared to the ground truth. We also experimented with a
`fact only' variant which only compares the facts, given and derived. The rationale
for this is that facts alone should provide a reliable proxy for the amount and quality of the reasoning
process. Derived facts are not present in the problem statement and for these to appear in
the response the model must have successfully carried out an inference. In contrast, rules could just be cited in a model response, without any reasoning. 

\subsection{Error Types}
\label{sec:MethodsErrorTypes}
We build on the error categories proposed by Xu et al. and propose a more specific system that could be used when evaluating LLMs on deductive reasoning \cite{Xu2023Large}. The  error categories described in Table \ref{tab:errortypes}  provide insight into the reasoning capabilities of the Large Language Models. Note that our `No Reasoning' error classification aligns perfectly with the corresponding category from Xu et al.'s classification. Our `Missing steps' category would roughly align with the `perspective mistake' and `process mistake' error types. Our `Wrong Method' category does not correspond to an error type per se, but instead is a measure of whether the model followed an ATP strategy as instructed. 

\begin{table}[!h]
  \centering
  \caption{Deductive Reasoning Error Types}
  \label{tab:errortypes}
  \begin{tabular}{c ccc}
    \hline
    Error Type & Description & Measurement Technique &  \\
    \hline
    \midrule
    \multirow{2}{*}{No Reasoning} & Where a model provides an  & Length of model's response &  \\
    & answer without reasoning & corresponds to `true' or `false' &  \\
    \hline
    \multirow{3}{*}{Missing Steps} & The model's answer is   & Determined using regular expressions  & \\
    &missing at least one of the   & on the model's reasoning and &\\
    &steps required for a proof  & converting to a consistent format  &\\
    \hline
    \multirow{3}{*}{Wrong Method} & The model does not  & Verified by determining if the order of &  \\
    & follow  the prescribed  & each of the statements matches what  &  \\
    & reasoning method &  would be expected by reasoning method &  \\
    \hline
    \hline
  \end{tabular}
\end{table}

\subsection{Natural Language Analysis}
\label{sec:natural-language-analysis}
As stated in the background section, Saparov and He \cite{Saparov2023Language} parse the
LLM statements with a recursive-descent parser using the simple grammar where unparseable
proofs are marked as incorrect. The method of using a strict grammar is likely to miss
many examples of natural language. To accept a wider variety of natural language
expressions we integrate into our workflow the open source natural language processing
library, spaCy \cite{Spacy}.  We use spaCy for shallow parsing the problem description,
ground truth reasoning, and model responses. We also used co-reference resolution, which
is frequently needed for correctly parsing the model response (we installed the 
`coreferee' extension library for that). SpaCy's built-in rule language was instrumental
for extracting facts and rules from the part-of-speech tagged documents. We wrote rules
that are conditioned on tags for named entities, lemmas and grammatical roles such as nouns,
adjectives and adverbs. The rules compute the facts and rules in the document expressed as
semantic triples over canonical word representations.

As an example take the following excerpt from an actual model response:
\begin{quote}
  \ttfamily Statement 5: lepidopterans are not hot.\\
  \text{...}\\
  Statement 8: lepidopteran are not sunny.\\
  Based on the analysis of the statements, we can
  conclude that Sally is a lepidopteran (statement 5), which means she is not hot and not
  sunny (statement 8).
\end{quote}
The result of the spaCy processing is the sequence of triples ``\texttt{lepidopteran is not hot.}'', 
 ``\texttt{lepidopteran is not sunny.}'', 
``\texttt{Sally is lepidopteran.}'', 
 ``\texttt{Sally is not hot.}''.

\subsection{Uncertainty}
Unlike some other areas of computer science which have algorithms with very consistent performance, LLM's performance is often random and can vary between similar tasks without good reason. For noisy scenarios like this, we believe that it is important to include uncertainty when reporting a model accuracy. Natural variation in measurements is commonplace in the physical sciences and the well established techniques from these fields \cite{GUM2008, Farrance2012Uncertainty}.

Taking repeat trails and the reporting of uncertainty in measurements is not always standard practice in all areas of computer science. We would like to argue for its importance in cases where variation naturally occurs in measurements. Consider the issue for readers who would like to generalise a newly published innovation for a task similar to the original published benchmarks. If only an accuracy is provided with no repeat trials or variation then they don't know how much variation they should when implementing this technique. Therefore we propose that a benchmark should be broken into multiple trials and then an average value and an uncertainty provided rather than just a single number.

As an example let's consider an example with two scenarios. In the first scenario, 1000 examples in a benchmark as split into 10 trials each which have 100 examples; so each example could be reported with an accuracy of $1\%$. Then the average could be calculated and the standard deviation across the 10 trials could be reported as the uncertainty or spread in the results. In the second scenario the are conducted for example accuracy could be reported to $0.1\%$ with no uncertainty values. We may think that the second approach provides more information as the accuracy value is more precise.

Our counter argument is this accuracy value would be exactly the same as the average accuracy calculated in the first scenario. However any reader or user who is interested in the results will not know if the model performance reported to $0.1\%$ accuracy us a `fluke' or whether they can rely on the model to consistently produce this result. 

We believe that this is incredibly important in case of leader-boards. For example is an accuracy of $89.2\%$ significantly better than an accuracy of $89.1\%$? This highly depends on the uncertainty; if both models  have a variation of approximately $5\%$ then the two techniques have no real difference in performance, one could have just been `lucky' that on this specific problem its accuracy was marginally higher. Uncertainty will help researchers to know if a new innovation is really valuable and making a significant impact over existing techniques. 

There are different techniques that could be used to calculate uncertainty in measurements. When there are relatively large numbers of repeat trials ($n=10+$) we recommend using the standard deviation as a good estimate of the uncertainty. In situations where there are a smaller number of repeat trials ($n=3-10$) we recommend using either half of the range. In situations where there are two values the entire range could be reported. Regardless of the method chosen, we believe it is important that authors make explicit their method for calculating uncertainty and the number of trials they conducted.  

In our experiments, each condition has three similar tasks (trials) of approximately $n=100$ problems. As the hosted and public models are regularly updated it was essential to conduct our experiments within a short time window to make sure that there were minimal updates to the model during the running of the experiment which could result in an unfair comparison of the different techniques. Therefore our results give a snapshot of the models' performance in December 2023. There were some cases where calls to the GPT and Gemini API's failed resulting in the loss of some model responses for some trials. Each trial contained a different number of steps in reasoning required to arrive at the answer, we vary from one hop to three hops in the PRONTOQA settings. We report the uncertainty as the half of the range between the three trials.

\section{Results}
\label{sec:results}

\subsection{Answer Correctness}

Appendix \ref{sec:preliminary} contains the results from the preliminary experiments that we performed to determine the best settings for the PRONTOQA dataset for our experiments. Table \ref{tab:accuracyresults} shows the accuracy for each of the experiments. The average number of completion tokens for each condition is given as a very rough measure of computational expense within each column. As expected the normal condition where models are expected to produce simply a True or False answer have the least completion tokens but also the lowest performance. The ATP reasoning strategies generally require a larger number of completion tokens, especially for GPT4 and Gemini. 

\begin{table}[!h]
  \centering
  \caption{Accuracy for False Ontology with Distractors Experiments. Average number completion tokens given in brackets. Random guessing would give an accuracy value of $0.500 \pm 0.042$.}
  \label{tab:accuracyresults}
  \begin{tabular}{>{\centering\arraybackslash}m{6cm}ccc}
    \hline
    Prompt Strategy & GPT3 & GPT4 & Gemini-Pro \\
    \hline
    \midrule
    \multirow{2}{*}{Normal} & $ 0.47\pm0.06 $ & $ 0.83\pm0.12 $ & $ 0.48\pm0.03 $ \\
    & $(19.7)$ & $(1.8)$ & $(3.0)$ \\
    \hline
    \multirow{2}{*}{}Explicit Instructions + & $ 0.65\pm0.1 $ & $ 0.75\pm0.09 $ & $ 0.62\pm0.09 $ \\
    Chain of Thought& $(461.5)$ & $(122.5)$ & $(287.7)$ \\
    \hline
    \multirow{2}{*}{}Explicit Instructions + One Shot + & $ 0.66\pm0.15 $ & $ 0.94\pm0.04 $ & $ 0.74\pm0.12 $ \\
    Chain of Thought& $(443.4)$ & $(92.3)$ & $(110.4)$ \\
    \hline
    \multirow{2}{*}{}LLM Reasoning Strategy & $ 0.69\pm0.02 $ & $ 0.950\pm0.015 $ & $ 0.72\pm0.03 $ \\
    Bottom Up& $(542.3)$ & $(311.2)$ & $(488.2)$ \\
    \hline
    \multirow{2}{*}{}LLM Reasoning Strategy & $ 0.56\pm0.05 $ & $ 0.987\pm0.005 $ & $ 0.56\pm0.06 $ \\
    Top Down& $(581.2)$ & $(250.31)$ & $(350.3)$ \\
    \hline
    \multirow{2}{*}{}LLM Reasoning Strategy & $ 0.64\pm0.07 $ & $ 0.943\pm0.015 $ & $ 0.77\pm0.02 $ \\
    Magic Set Transformation& $(366.4)$ & $(371.4)$ & $(402.2)$ \\
    \hline
  \end{tabular}
\end{table}

The results are overall consistent with other studies, that model performance is improved with explicit instructions, an example and CoT reasoning. At the time of experiment (December 2023), GPT4 was the state of the art model and as expected its performance is significantly higher than GPT3 and Gemini-Pro. At first glance it appears that explicit instructions and CoT reasoning alone has decreased the performance of GPT4, however the high uncertainty values show that this finding is not significant and that the performance is comparable within uncertainty. Note that as accuracy improves there is a general trend that the uncertainty decreases. 

\subsection{Process Hard Correctness}
\begin{figure}[!h]
    \centering
\begin{tikzpicture}
\begin{axis}[ybar,
        bar width=7pt,
        width=0.9\textwidth,
        title={Percentage of Cases where each Model Displayed Reasoning by Technique},
        legend style={at={(0.97,-0.15)}, anchor=north east, legend columns=-1, font=\scriptsize, /tikz/column sep=5pt},
        ymin=0,
        ymax=100,        
        ylabel={\small{\% of cases including some reasoning}},
        xtick pos=bottom,
        xtick=data, 
        xticklabels = {
            GPT3,
            GPT4,
            Gemini           
        },
        major x tick style = {opacity=0},
        minor x tick num = 1,
        minor tick length=2ex,        
        xticklabel style={text width=1.5cm, align=center}, 
        enlarge x limits=0.3
        ]
\addplot+[error bars/.cd, y dir=both, y explicit, error bar style={black}]
    table[x expr=\coordindex, y=Y, y error plus=PosErr, y error minus=NegErr] {\datatableisreasoningZero};
\addplot+[error bars/.cd, y dir=both, y explicit, error bar style={black}]
    table[x expr=\coordindex, y=Y, y error plus=PosErr, y error minus=NegErr] {\datatableisreasoningOne};
\addplot+[error bars/.cd, y dir=both, y explicit, error bar style={black}]
    table[x expr=\coordindex, y=Y, y error plus=PosErr, y error minus=NegErr] {\datatableisreasoningTwo};
\addplot+[error bars/.cd, y dir=both, y explicit, error bar style={black}]
    table[x expr=\coordindex, y=Y, y error plus=PosErr, y error minus=NegErr] {\datatableisreasoningThree};
\addplot+[error bars/.cd, y dir=both, y explicit, error bar style={black}]
    table[x expr=\coordindex, y=Y, y error plus=PosErr, y error minus=NegErr] {\datatableisreasoningFour};
\addplot+[error bars/.cd, y dir=both, y explicit, error bar style={black}]
    table[x expr=\coordindex, y=Y, y error plus=PosErr, y error minus=NegErr] {\datatableisreasoningFive};

\end{axis}
\end{tikzpicture}

\begin{tikzpicture}
\begin{axis}[ybar,
        bar width=7pt,
        width=0.9\textwidth,
        title={Percentage of Cases where each Model Displayed All Required Steps},
        legend style={at={(0.97,-0.15)}, anchor=north east, legend columns=-1, font=\scriptsize, /tikz/column sep=5pt},
        ymin=0,
        ymax=100,        
        ylabel={\small{\% of cases including all reasoning steps}},
        xtick pos=bottom,
        xtick=data, 
        xticklabels = {
            GPT3,
            GPT4,
            Gemini           
        },
        major x tick style = {opacity=0},
        minor x tick num = 1,
        minor tick length=2ex,        
        xticklabel style={text width=1.5cm, align=center}, 
        enlarge x limits=0.3
        ]
\addplot+[error bars/.cd, y dir=both, y explicit, error bar style={black}]
    table[x expr=\coordindex, y=Y, y error plus=PosErr, y error minus=NegErr] {\datatablehasstepsZero};
\addplot+[error bars/.cd, y dir=both, y explicit, error bar style={black}]
    table[x expr=\coordindex, y=Y, y error plus=PosErr, y error minus=NegErr] {\datatablehasstepsOne};
\addplot+[error bars/.cd, y dir=both, y explicit, error bar style={black}]
    table[x expr=\coordindex, y=Y, y error plus=PosErr, y error minus=NegErr] {\datatablehasstepsTwo};
\addplot+[error bars/.cd, y dir=both, y explicit, error bar style={black}]
    table[x expr=\coordindex, y=Y, y error plus=PosErr, y error minus=NegErr] {\datatablehasstepsThree};
\addplot+[error bars/.cd, y dir=both, y explicit, error bar style={black}]
    table[x expr=\coordindex, y=Y, y error plus=PosErr, y error minus=NegErr] {\datatablehasstepsFour};
\addplot+[error bars/.cd, y dir=both, y explicit, error bar style={black}]
    table[x expr=\coordindex, y=Y, y error plus=PosErr, y error minus=NegErr] {\datatablehasstepsFive};

\legend{Normal, CoT, CoT one-shot, Bottom up, Top down, Magic set}
\end{axis}
\end{tikzpicture}
    \caption{Presence of Reasoning. The top graph shows the percentage of cases where each model showed some form of reasoning to solve the problem. The bottom graph shows the percentage of cases where the models displayed all of the steps that an ATP would go through to solve the problem.}
    \label{fig:reasoningpresent}
\end{figure}
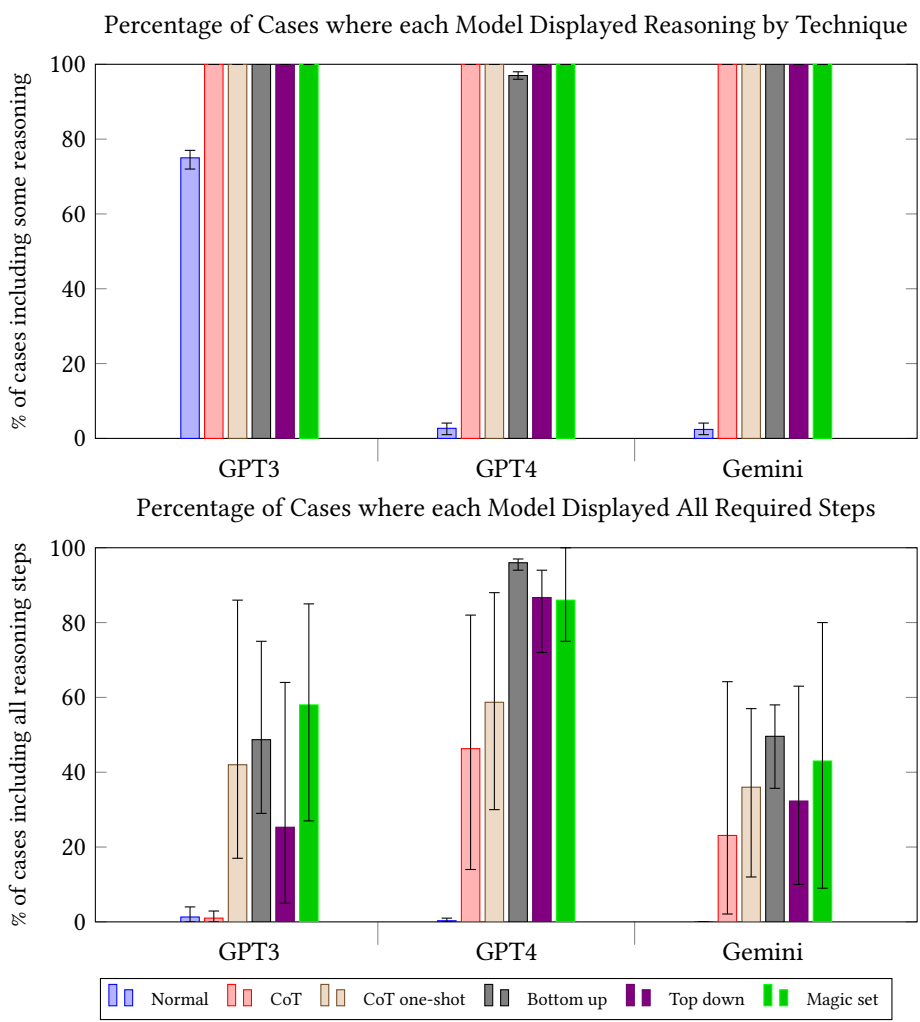

Figure \ref{fig:reasoningpresent} contains two distinct graphs. The first graph illustrates the portion of cases where the model included some reasoning. For all conditions except `normal', all models consistently included some reasoning, even if it was not correct. In the normal condition models were not asked to show working and only GPT3 regularly included some justification/reasoning for its answer.

The second graph in Figure \ref{fig:reasoningpresent} investigates the number of cases where the model included all of the steps of reasoning that would be required to construct a complete proof. There is a significant variation between different trials for all experimental conditions. On average GPT4 was most likely to include all required reasoning steps, especially when given explicit instructions for bottom up reasoning.

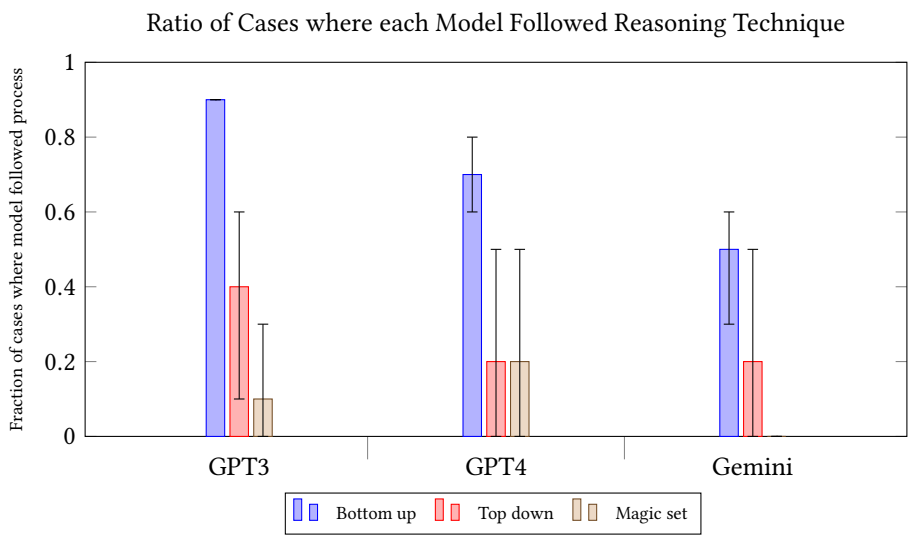
\begin{figure}[!h]
    \centering
\begin{tikzpicture}
\begin{axis}[ybar,
        bar width=7pt,
        width=0.9\textwidth,
        title={Ratio of Cases where each Model Followed Reasoning Technique},
        legend style={at={(0.75,-0.15)}, anchor=north east, legend columns=-1, font=\scriptsize, /tikz/column sep=5pt},
        ymin=0,
        ymax=1,        
        ylabel={\scriptsize{Fraction of cases where model followed process}},
        xtick pos=bottom,
        xtick=data, 
        xticklabels = {
            GPT3,
            GPT4,
            Gemini           
        },
        major x tick style = {opacity=0},
        minor x tick num = 1,
        minor tick length=2ex,        
        xticklabel style={text width=1.5cm, align=center}, 
        enlarge x limits=0.3
        ]
\addplot+[error bars/.cd, y dir=both, y explicit, error bar style={black}]
    table[x expr=\coordindex, y=Y, y error plus=PosErr, y error minus=NegErr] {\datatableratiorightmethodandhasstepsThree};
\addplot+[error bars/.cd, y dir=both, y explicit, error bar style={black}]
    table[x expr=\coordindex, y=Y, y error plus=PosErr, y error minus=NegErr] {\datatableratiorightmethodandhasstepsFour};
\addplot+[error bars/.cd, y dir=both, y explicit, error bar style={black}]
    table[x expr=\coordindex, y=Y, y error plus=PosErr, y error minus=NegErr] {\datatableratiorightmethodandhasstepsFive};

\legend{Bottom up, Top down, Magic set}
\end{axis}
\end{tikzpicture}
    \caption{Ratio of Process Following. This graph shows the ratio of times the model followed the instructed reasoning process out of all attempts where each model showed all required reasoning steps.}
    \label{fig:correctprocess}
\end{figure}

Figure \ref{fig:correctprocess} displays results relating specifically to the experiments where the models were given a specific reasoning technique to follow. It shows the number of cases where the model followed the specified reasoning process as a fraction of the cases that contained all reasoning steps. The results show that all models were most able to follow the bottom up reasoning process. 

\begin{figure}[!h]
    \centering
    \begin{subfigure}{.341\textwidth}
  \centering
  \includegraphics[width=\linewidth]{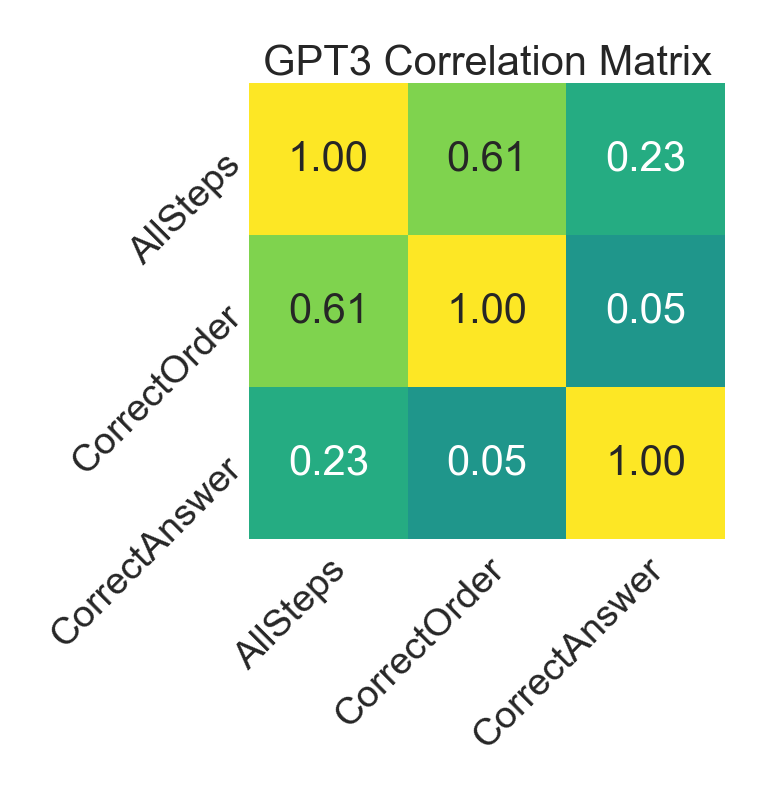}
  \caption{GPT-3 Correlation Matrix}
  \label{fig:sub1}
\end{subfigure}%
\hfill
\begin{subfigure}{.26\textwidth}
  \centering
  \includegraphics[width=\linewidth]{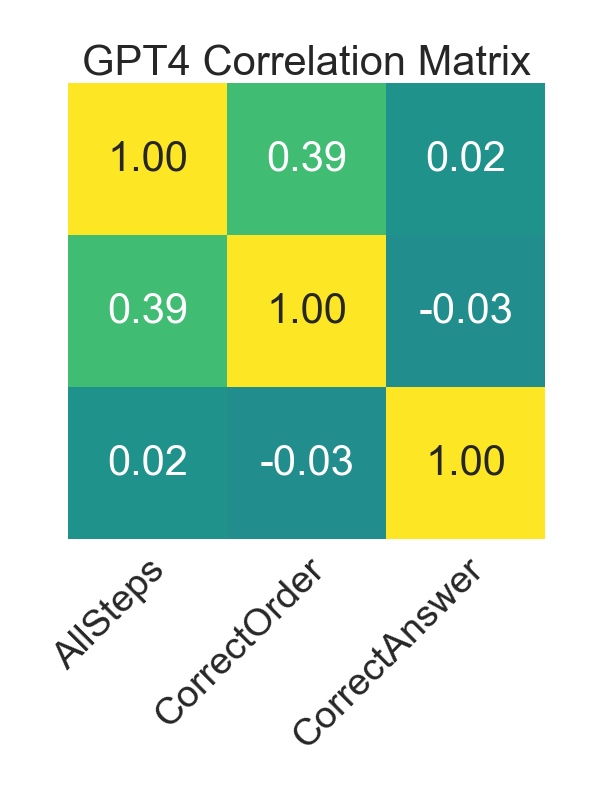}
  \caption{GPT-4 Correlation Matrix}
  \label{fig:sub2}
\end{subfigure}%
\hfill
\begin{subfigure}{.332\textwidth}
  \centering
  \includegraphics[width=\linewidth]{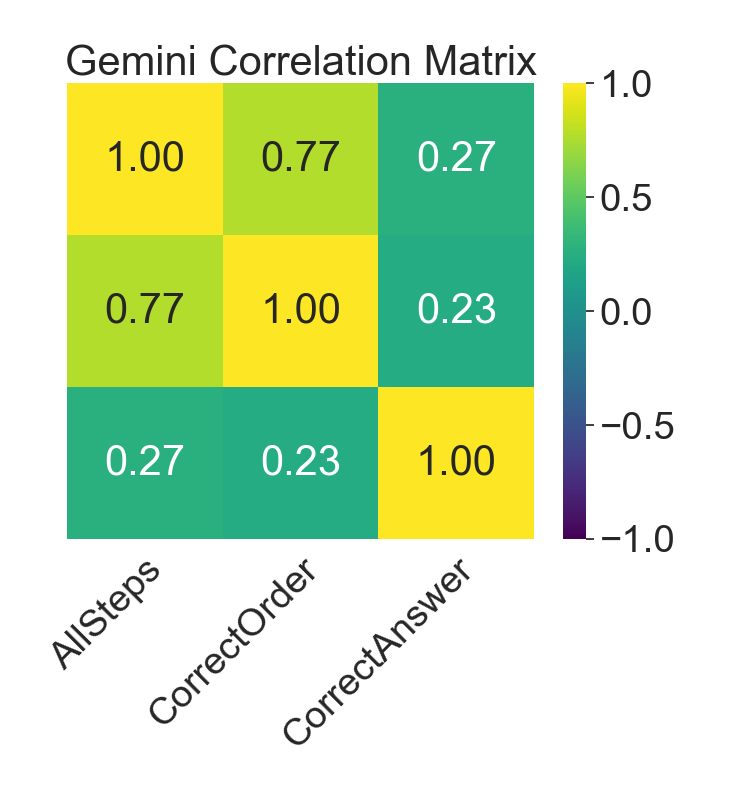}
  \caption{Gemini Correlation Matrix}
  \label{fig:sub3}
\end{subfigure}
    \caption{Correlation Matrices. These matrices shows the correlation between whether the model included all steps (AllSteps), included the steps in the correct order (CorrectOrder) and obtain the correct answer (CorrectAnswer).}
    \label{fig:CorMat}
\end{figure}

Figure \ref{fig:CorMat} contains three correlation matrices for each of the three models, GPT3, GPT4 and Gemini. We combine the bottom up, top down and magic set results. We then calculate the correlation between inclusion of all required steps in order and obtaining the correct answer. Note that obtaining all of the steps is a precursor to having these steps in the correct order and therefore there is a strong positive correlation between `All Steps' and `Correct Answer'. 

\subsection{Process Soft Correctness}

As introduced in Section~\ref{sec:methods}, process soft correctness relates the facts and
rules given as ground truth ($\mathit{gt}$) with those found in the model response ($\mathit{mr}$).
These sets are found by natural language analyses as explained in Section~\ref{sec:natural-language-analysis}.
We define measures of recall and precision as follows:
\begin{xalignat*}{2}
 \text{recall}(\mathit{gt}, \mathit{mr}) & = |\mathit{gt} \cap \mathit{mr}| / |\mathit{gt}|
  &
 \text{precision}(\mathit{gt}, \mathit{mr}) & = |\mathit{gt} \cap \mathit{mr}| / |\mathit{mr}|
\end{xalignat*}
With \emph{recall} we measure the degree of the ground truth (completeness) discovered in the model
response, and with \emph{precision} we measure a degree of redundancy (inverse of conciseness) of the facts and
rules in the model response compared to the ground truth.

We distinguish two versions of building the sets $\mathit{gt}$ and $\mathit{mr}$.
The \emph{full} version comprises both facts and rules, as explained above, whereas in
the \emph{facts only} version only the facts are retained. As indicated in
Section~\ref{sec:methods}, the rationale is that facts by themselves are a sharper
proxy for assessing the reasoning carried out by a model.

We computed recall and precision statistics for all models and techniques except the
`normal' technique because the models are not expected to produce facts and rules for this condition.
We filtered out runs with $\mathit{mr} = \emptyset$ because recall and precision is meaningless in this
case. We found that the results are similar for the full
and the facts only versions for $\mathit{gt}$ and $\mathit{mr}$.
However, as expected, they are somewhat more pronounced in the facts only version and we 
report on these results only.

We investigated how sensitive the statistics are to certain parameter settings that we
thought could have an impact. We distinguished fixing `number of hops' to 1, 2 or 3; whether
the query is negated (``\texttt{True or false: Sally is not hot.}'') or not; and whether the
correct answer is `true' or is `false'. We found that these parameters have little 
impact on the statistics even in combination.
A parameter that we found significant, though, is whether the question was answered
correctly or not.

With these considerations and investigations we collected results for all runs
classified by model, technique, and whether a run correctly answered the question.
Table~\ref{tab:soft} provides a summary. Notice that recall and precision
are stated as the mean and standard deviation over all runs in a class.
\begin{table}[htpb]
  \centering
{\small
\renewcommand{\arraystretch}{1.1}
\begin{tabular}{llRRRRRR}
 &   & \multicolumn{3}{c}{\bfseries Correctly answered} & \multicolumn{3}{c}{\bfseries Incorrectly answered} \\
    \multicolumn{1}{l}{\bfseries Model} &
    \multicolumn{1}{l}{\bfseries Technique} &
    \multicolumn{1}{c}{n~(\%)} & \multicolumn{1}{c}{Recall} & \multicolumn{1}{c}{Precision} &
        \multicolumn{1}{c}{n~(\%)} & \multicolumn{1}{c}{Recall} & \multicolumn{1}{c}{Precision}\\\hline\hline
GPT3 & BottomUp & 201~(67\%) & 0.80 \pm 0.26 & 0.41 \pm 0.19 & 98~(33\%) & 0.66 \pm 0.28 & 0.33 \pm 0.19\\
 & CoT1Shot & 197~(66\%) & 0.76 \pm 0.26 & 0.51 \pm 0.20 & 102~(34\%) & 0.64 \pm 0.23 & 0.45 \pm 0.16\\
 & CoT & 181~(93\%) & 0.02 \pm 0.11 & 0.10 \pm 0.17 & 13~(7\%) & 0.08 \pm 0.28 & 0.50 \pm 0.00\\
 & MagicSet & 185~(62\%) & 0.87 \pm 0.21 & 0.50 \pm 0.17 & 114~(38\%) & 0.75 \pm 0.24 & 0.38 \pm 0.17\\
 & TopDown & 166~(56\%) & 0.66 \pm 0.27 & 0.51 \pm 0.19 & 133~(44\%) & 0.61 \pm 0.26 & 0.43 \pm 0.16\\
\hline
GPT4 & BottomUp & 278~(96\%) & 0.98 \pm 0.11 & 0.43 \pm 0.23 & 12~(4\%) & 0.97 \pm 0.10 & 0.33 \pm 0.25\\
 & CoT1Shot & 281~(94\%) & 0.90 \pm 0.19 & 0.71 \pm 0.22 & 18~(6\%) & 0.67 \pm 0.28 & 0.57 \pm 0.25\\
 & CoT & 221~(74\%) & 0.78 \pm 0.26 & 0.71 \pm 0.25 & 78~(26\%) & 0.67 \pm 0.32 & 0.59 \pm 0.29\\
 & MagicSet & 282~(94\%) & 0.95 \pm 0.14 & 0.55 \pm 0.16 & 17~(6\%) & 0.98 \pm 0.08 & 0.43 \pm 0.19\\
 & TopDown & 295~(99\%) & 0.93 \pm 0.19 & 0.62 \pm 0.17 & 4~(1\%) & 0.81 \pm 0.38 & 0.40 \pm 0.19\\
  \hline
Gemini-Pro & BottomUp & 214~(72\%) & 0.75 \pm 0.34 & 0.45 \pm 0.22 & 83~(28\%) & 0.62 \pm 0.30 & 0.38 \pm 0.17\\
 & CoT1Shot & 221~(74\%) & 0.75 \pm 0.29 & 0.63 \pm 0.28 & 78~(26\%) & 0.39 \pm 0.30 & 0.33 \pm 0.27\\
 & CoT & 181~(63\%) & 0.41 \pm 0.43 & 0.52 \pm 0.25 & 108~(37\%) & 0.29 \pm 0.38 & 0.47 \pm 0.24\\
 & MagicSet & 231~(77\%) & 0.82 \pm 0.22 & 0.60 \pm 0.17 & 68~(23\%) & 0.58 \pm 0.27 & 0.36 \pm 0.15\\
 & TopDown & 167~(56\%) & 0.69 \pm 0.28 & 0.48 \pm 0.20 & 132~(44\%) & 0.61 \pm 0.28 & 0.42 \pm 0.16\\
  \hline
\end{tabular}
}
\caption{Recall and precision analysis for process soft correctness discriminating whether
  the question was answered correctly or not. 
Recall and precision values are of the form $\mathit{Mean} \pm \mathit{StdDev}$. The columns $n (\%)$ show the counts and percentages of data points in a class.
}
  \label{tab:soft}
\end{table}

\section{Discussion}
\label{sec:discussion}
The overall accuracy results (Table \ref{tab:accuracyresults}) show that teaching models to use explicit reasoning strategies produces a performance which is comparable with on-shot CoT reasoning within uncertainty. Although the accuracy is similar between explicit reasoning strategies and CoT, the completion tokens (and therefore computation expense) is more than $200\%$ higher for all explicit reasoning strategies when used by GPT4 and Gemini. Both these completion token values and Figure \ref{fig:reasoningpresent} show that GPT4 and Gemini are very concise and rarely give reasoning when not asked. GPT3 on the other hand was more likely to provide justification even when this was not requested. 

Although models would consistently produce some reasoning (or `working') when asked, Figure \ref{fig:reasoningpresent} shows that this reasoning did not always contain all of the required steps. This can occur because the model skips steps or because the model was unable to complete a reasoning process and arrive at an answer. We note that whether the model was able to include all reasoning steps or not was largely independent of the type of prompt used, excepting the normal condition. GPT3 was consistently unable to produce all of the required reasoning steps without at least one example in the prompt.

Consistent with previous literature, we show that models have a preference for following a bottom up style reasoning procedure \cite{Kazemi2023LAMBADA}. At a high level, the bottom up approach works well for LLMs because  the intermediate steps involve more writing of more facts. This essentially provides place for the model to think. Top down reasoning is particularly difficult for LLMs because it involves backtracking and keeping track of which paths have already been explored. The magic set approach is also more difficult for LLMs because it has more steps in the process where the models could make a mistakes. 

One very interesting finding is the lack of a correlation between including all of the required reasoning to prove the final answer and then obtaining the correct answer itself. Figure \ref{fig:CorMat} shows that the correlation between having the correct reasoning and the correct answer is less than $0.3$ in all cases and close to $0$ in half of the cases. This result is significant and has significant implications for the trustworthiness of LLMs as it shows that even if a Large Language Model can demonstrate all of the correct reasoning it still may not obtain the correct answer. It also shows that a Large Language model is able to obtain the correct answer even in cases where it skips steps in the working/proof.

Regarding process soft correctness, the results in Table~\ref{tab:soft} indicate a correlation between recall and correctly answering a problem.
Across almost all classes, recall is higher for correctly answered questions than for
incorrectly answered questions. (The biggest difference is for `Gemini-Pro' and `MagicSet',
with recall values of 0.82 vs.\ 0.58.) This allows us to conclude that correct answers
tend to be produced along with the facts relevant for deriving the result. 
\emph{However, the converse is not generally true, that
instructing a model to derive relevant facts helps in computing a correct answer.}
This can be seen by observing that the strongest guided techniques, `MagicSet' and
`TopDown' do not consistently produce a significantly higher share of correct answers than the
non-guided techniques.
Nevertheless it is interesting to see that `MagicSet' delivers comparatively high recall and high
precision rates. This suggests, as future work, to consider a hybrid architecture where
techniques like `MagicSet' could be used to extract a small set of relevant facts that can
be handed off to a trusted inference engine for actually computing a correct answer.

One limitation of this study is that although performance measures such as accuracy can be made for each model, the computational expense cannot be compared between models. For models that are run locally, there are accurate measures of computational expense such as runtime/CPU time/GPU time. If models are not run locally, but the number of parameters of each model is known then it would be possible to multiply completion tokens by the number of parameters as a rough estimate of computational expense. This was not possible in this study as the number of parameters of the GPT4 and Gemini models is not published.

\section{Conclusions and Future Work}
\label{sec:conclusions}
This research is the first to determine the ability of LLMs to use ATP reasoning strategies. We evaluate the performance of three of the best available models on steamroller problems using the most difficult settings of the PRONTOQA Benchmark. In terms of accuracy the models' performance when using the ATP reasoning strategies was similar to one-shot CoT, despite the computational expense of executing the bottom up, top down and magic set reasoning being significantly higher than the model's natural reasoning. We found that LLMs have a preference for using bottom up processes.

We build on previous studies which have used  recursive-descent parsers by introducing the existing robust Natural Language Processing library, spaCy to this domain of study. SpaCy provides useful out of the box functionality for co-referencing, named entity recognition, point of speech tagging and a built in rule language. This enabled the reliable translation of the results into a consistent format for parsing and and evaluation. The use of these NLP toos allowed us to scalably dissect and evaluate the models' reasoning processes. 

This allowed us to discover that models were able to `skip steps' and obtain correct answers without showing all of the stepts required for a proof. We also found that models were able to include all of the steps required for a proof in the correct order, but still give the incorrect answer to a question. Overall, there was little correlation between correct reasoning and correct answers for all three of the models. This has a negative implication for the explainability of LLMs, even with current state of the art models, their reasoning cannot be trusted.

In terms of overall model comparison, GPT4 had systematically higher accuracy than either GPT3 and Gemini. It was also able to more consistently generate correct reasoning and include all required steps for a logical proof. Gemini and GPT4 were also found to be more concise than GPT3; on average they used less completion tokens for a given prompt. We were unable to compare the computational expense of different models as key information such as the number of parameters of each model, was not available. 

As area for future work we propose to investigate how LLMs reason on tasks with a greater number of rules and facts. As research in ATP has shown, it is only when problems reach a certain size that the choice of reasoning strategy becomes relevant. It is unclear where such a threshold would be for LLM reasoning. There may be a point where the benefits of top down reasoning approaches outweigh the model's natural tendency to perform bottom up reasoning. 

One area that also deserves more exploration in this field is the use of neuro-symbolic approaches which combine LLMs with ATP systems. Traditionally ATP systems have been limited by their rigidity and the need to manually input information. If an LLM is able to relatively accurately accomplish this manual labour then the logical reasoning could be `outsourced' to the ATP systems allowing for excellent explainability and interpretability. This naturally lends itself to problems where justifications and explanations are required such as education, healthcare, law and banking.


\bibliographystyle{ACM-Reference-Format}
\bibliography{SteamRollerProblems}

\appendix

\clearpage

\section{Appendix - Preliminary Experiments}
\label{sec:preliminary}

Table \ref{Tab:preliminary} contains results from preliminary experiments in order to determine the most appropriate PRONTOQA parameter settings for the experiments. For experimental conditions without distractors we tested with one to five hops (therefore five trials); experimental conditions with distractors have one to three hops (three trials). One hundred problems were included in each trial. 

\begin{table}[!h]
  \centering
  \caption{Comparison of Random Guessing, GPT-3 and GPT-4 under Various Experimental Conditions. No Chain of thought or one/few shot prompts were used.}
  \label{Tab:preliminary}
  \begin{tabular}{|>{\centering\arraybackslash}m{6cm}|c|c|}
    \hline
     \multirow{2}{*}{Experimental Condition} & \multirow{2}{*}{ Accuracy} & Average Completion  \\
    &   &  and Prompt Tokens \\
    \hline
    Random Guessing & $0.50\pm0.042$ & $(0, 0)$ \\
    \hline
    GPT-3 True Ontology & $0.994\pm0.005$ & $(20 , 44.9)$ \\
    \hline
    GPT-3 False Ontology & $0.686\pm0.075$ & $(25.3 , 41.6)$ \\
    \hline
    GPT-3 Fictional Ontology & $0.902\pm0.04$ & $(27.6 , 45.7)$ \\
    \hline
    GPT-3 False Ontology with Distractors & $0.487\pm0.06$ & $(19.7,110.5)$ \\
    \hline
    GPT-4 False Ontology & $0.974\pm0.015$ & $(3.68 , 41.6)$ \\
    \hline
    GPT-4 False Ontology with Distractors & $0.833\pm0.115$ & $(1.8,110.5)$ \\
    \hline
  \end{tabular}
  
\end{table}

This preliminary experiment shows that even the weaker model GPT3, shows ceiling effects on the tasks with the True ontology. To best prevent ceiling effects with the stronger GPT4 model, we chose to use the most difficult condition, False Ontology with Distractors. We would like to point out that a merit of the fictional ontology is that there is less chance of the model being able to use external knowledge to solve the problem rather than going through a reasoning process. However if contamination has occurred, the fictional ontology is likely to be more vulnerable because it contains combinations of tokens that would only be seen in this data set. 

We also performed an experiment to determine the suitability of using different numbers of hops as repeat trials. In Table \ref{Tab:hops} we present the results. For problems without distractors changing the number of hops makes causes very little variation in the model performance. We attribute this to the same cause as the authors of the dataset \cite{Saparov2023Language}, who claim that in these cases the model can avoid the task of reasoning by instead performing an alternative task with the same result. When there are no distractors the answer can easily be obtained by simply counting the number of times `not' appears in the problem statement. 

\begin{table}[!h]
  \centering
  \caption{Model accuracy, completion tokens and prompt tokens `(completion tokens, prompt tokens)' on Various Experimental Conditions Across Different Number of Hops. The right hand column shows the range in percentage accuracy among the different number of hops. Prompt engineering techniques were used. The overall pattern is that there is very little change in accuracy with number of hops until distractors are added. As expected the number of prompt tokens steadily increases with number of hops.}
  \label{Tab:hops}
  \small
  \begin{tabular}{|>{\centering\arraybackslash}m{2cm}|c|c|c|c|c|c|}
    \hline
    Experimental Conditions & 1 Hop & 2 Hops & 3 Hops & 4 Hops & 5 Hops & Range \\
    \hline
    \multirow{2}{*}{ } GPT3 True Ontology & $0.99$ & $0.99$ & $1$ & $0.99$ & $1$ & \multirow{2}{*}{$1\% $}  \\
    & $(9.9 , 30.4)$ & $(14.3 , 37.64)$ & $(22.2 , 45.2)$ & $(22.8 , 51.7)$ & $(30.8 , 59.6)$ & \\
    \hline
    \multirow{2}{*}{ } GPT3 False Ontology & $0.76$ & $0.79$ & $0.79$ & $0.81$ & $0.73$ & \multirow{2}{*}{$8\%$} \\
    & $(23.1, 27.5)$ & $(24.2 , 35.0)$ & $(25.3 , 41.7)$ & $(26.1 , 48.8)$ & $(27.8 , 54.9)$ & \\
    \hline
    \multirow{2}{*}{} GPT3 Fictional Ontology & $0.91$ & $0.95$ & $0.90$ & $0.87$ & $0.88$ & \multirow{2}{*}{$7\%$} \\
    & $(13.0 , 29.2)$ & $(23.4 , 37.2)$ & $(33.2 , 46.9)$ & $(33.8 , 53.5)$ & $(34.6 , 61.98)$ & \\
    \hline
    \multirow{2}{*}{ } GPT3 False Ontology with Distractors & $0.53$ & $0.52$ & $0.41$ & N/A & N/A & \multirow{2}{*}{$12\%$} \\
    & $(16.5 , 95.8)$ & $(21.7 , 108.9)$ & $(20.9 , 123.7)$ &  &  & \\
    \hline
    \multirow{2}{*}{ } GPT4 False Ontology & $0.96$ & $0.97$ & $0.99$ & $0.97$ & $0.98$ & \multirow{2}{*}{$3\%$} \\
    & $(4, 27.5)$ & $(3.36 , 35.0)$ & $(1.89 , 41.7)$ & $(3.17, 48.8)$ & $(5.99 , 54.9)$ & \\
    \hline
    \multirow{2}{*}{ } GPT4 False Ontology with Distractors & $0.94$ & $0.85$ & $0.71$ & N/A & N/A & \multirow{2}{*}{$23\%$} \\
    & $(2.04 , 95.8)$ & $(1.17 , 108.9)$ & $(2.19 , 123.7)$ &  &  & \\
    \hline
  \end{tabular}
  \normalsize
\end{table}

\section{Appendix - Prompts}
\label{sec:prompts}

For each of the prompt techniques shown below three sets (1 hop, 2 hops and 3 hops) of 100 questions and queries were provided. An example question and query are provided below:

\begin{quote}
    \{question\} = \textit{``Each sheep is sunny. Each sheep is a feline. Sheep are mammals. Felines are aggressive. Every feline is a snake. Felines are carnivores. Each snake is luminous. Snakes are cats. Every dog is not luminous. Each snake is an animal. Animals are fast. Carnivores are opaque. Each mammal is floral. Each vertebrate is not feisty. Each vertebrate is a cow. Alex is a sheep. Alex is a vertebrate.''}
\end{quote}

\begin{quote}
    \{query\} = \textit{``True or false: Alex is luminous.''}
\end{quote}

The prompt for the normal condition was as follows - Average number of prompt tokens is $110.5$ for GPT3 and GPT 4 and $94.2$ for Gemini:
\begin{quote}
    prompt = \textit{``\{question\} \{query\}''}
\end{quote}

\paragraph{Chain of thought prompt} - Average number of prompt tokens is $162.3$ for GPT and $148$ for Gemini:
\begin{quote}
    prompt = \textit{``Consider the following statements and the given query. Use your reasoning skills to determine if the query is true or false based on the statements. Explain your thought process step by step as you analyze the relationship between the statements and the query. \\
    Statements: \{question\} \\
    Query: \{query\}''}
\end{quote}

Chain of thought and one-shot - Average number of prompt tokens is $301.3$ for GPT and $283$ for Gemini:
\begin{quote}
    prompt = \textit{``Consider the following statements and the given query. Use your reasoning skills to determine if the query is true or false based on the statements. Follow the format of the example question that follows.\\
    Example Statements: `All cats are birds. No bird swims. Whiskers is a cat.' \\
    Query: `True or false: Whiskers swims.' \\
    Example Reasoning: `Let's figure out if Whiskers swims. This is not provided directly in the statements. However it does state that Whiskers is a cat. Then it states that all cats are birds. Therefore Whiskers is a bird. It then states that no bird swims. Since Whiskers is a bird this means that Whiskers does not swim. Therefore the query ``Whiskers Swims'' is false.' \\
    Explain your thought process step by step as you analyze the relationship between the statements and the query. \\
    Statements: \{question\} \\
    Query: \{query\}''}
\end{quote}

Bottom up prompt - Average number of prompt tokens is $510.3$ for GPT and $496.7$ for Gemini:
\begin{quote}
    prompt = \textit{``Consider the following statements, which include rules and then facts, along with the given query. Facts are have a specific named instance. For example ``Sam is a cat'' is a fact because an explicit name ``Sam'' is given. Rules establish a connection between two general classes without referring to a specific instance. For example ``Cats are birds'' is rule, not a fact because there is no named instance of a specific cat or bird. Modus ponens can be used to apply a rule to a fact. For example applying ``Cats are birds'' to ``Sam is a cat'' gives a new fact ``Sam is a bird''. To use bottom-up reasoning: \\
        - List the rules which do not contain specific named instances.\\
        - List all given facts with a specific named instance as a current facts. \\
        - Then for each rule do the following:\\
        1.) Examine each rule to see if its premise applies to any of the current facts. It's important to recognize that general rules can apply to specific facts. For example, the rule 'All cats are birds' applies to the fact `Sam is a cat,' even though the rule doesn't mention Sam specifically. This is because the rule involves a category ('cats') that matches a variable in the fact ('cat'). So, when checking for a match, look beyond direct mentions and consider whether the rule's general premise encompasses the specifics of the current facts. This careful matching is essential to correctly apply rules to facts. Then add the new concluded fact (in this example ``Sam is a bird'') to the list of current facts.\\
        2.) Check if the current facts contain the query or its negation. If they do, answer the query. If it does not, repeat this procedure for all rules again.
        Iterate through this process until a conclusion is reached. It may take three or four iterations to reach a conclusion that will answer the query.\\
        Statements: \{question \} \\
        Query:\{ query \}''}
\end{quote}

Top down prompt - Average number of prompt tokens is $499.3$ for GPT and $485$ for Gemini::
\begin{quote}
    prompt = \textit{``Consider the following statements, which include rules and then facts, along with the given query. Use a top-down reasoning strategy to answer the query. Facts have a specific named instance. For example, `Sam is a cat' is a fact because an explicit name `Sam' is given. Rules establish a connection between two general classes without referring to a specific instance. For example, `Cats are birds' is a rule, not a fact, because it does not name a specific cat or bird. To use top-down reasoning, follow these steps:\\
    1. List the rules which do not contain specific named instances.\\
    2. List all given facts with a specific named instance as current facts.\\
    3. Then consider the query and repeat the following procedure:\\
    4. Check if the query or its negation is among the facts. If so, answer the query.\\
    5. If it is not, then search through the set of rules and check if the conclusion of any of the rules matches the query. If so, make the body of this rule the new query, paying special attention to maintaining the correct use of negation throughout the process. For example, if your query is `Does Sam not swim?' and you have a rule like `Birds swim', you would update the query to `Is Sam not a bird?'. Re-write the query with this new perspective and return to step 4.\\
    You will need to repeat the procedure multiple times until the query or its negation appear as facts. It is crucial to meticulously track all instances of 'not' or other negations when updating the query, as this directly affects the final answer.\\
    It may take three or four iterations to arrive at the final answer. If a particular line of reasoning comes to a dead end, it might help to start again with the original query and try applying different rules.\\
Statements: \{question\}\\
Query: \{query\}''}
\end{quote}

Magic set prompt - Average number of prompt tokens is $444.3$ for GPT and $431$ for Gemini::
\begin{quote}
    prompt = \textit{``Consider the following statements, which include both rules and facts, along with the given query. Approach the query with a top-down reasoning strategy, keeping in mind that facts are specific instances like `Sam is a cat,' and rules are general principles like `Cats are birds.' Execute the following steps:\\
1. List the general rules and specific named facts.\\
2. Create a set of subgoals starting with the query, aiming to gradually uncover the truth of the query through logical deduction.\\
3. Expand subgoals by systematically adding the premises of rules that directly relate to these subgoals. Continue this process until no new subgoals emerge.\\
4. Refine your set of rules by eliminating any that don't contribute to achieving the subgoals.\\
5. Carefully match the premises of the refined rules with the listed facts. Consider how general rules apply to specific instances and derive any new facts from this matching. Be particularly attentive to the implications of negations and complex statements.\\
6. Check if the current set of facts directly resolves the query or its negation. If a resolution is found, provide the answer.\\
7. If the answer is not yet clear, reapply the refined rules and facts iteratively, with each iteration aiming to reveal new insights or facts that help answer the query. In each iteration, aim to either refine your understanding or deduce a new fact that edges closer to resolving the query.\\
Continue this iterative process of application and review until a definitive conclusion is reached regarding the query. The process demands careful analysis and might require several iterations, so remain diligent and focused on how each piece of information contributes to your understanding of the query.\\
Statements: \{question\}\\
Query: \{query\}''}
\end{quote}

\end{document}
\endinput